# A serial dual-channel library occupancy detection system based on Faster RCNN

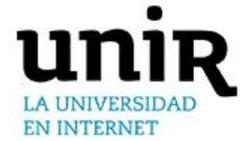

Guoqiang Yang[1, †], Xiaowen Chang[2, †], Zitong Wang[3, †], Min Yang[1, *]

[1] School of Space Science and Technology, Xidian University, Xi'an (China)
[2] SDU-ANU Joint Science College, Shandong University, Jinan (China)
[3] School of Software Engineering, South China University of Technology, Guangzhou (China)
[†] These authors contributed equally to this work.

## ABSTRACT

The phenomenon of seat occupancy in university libraries is a prevalent issue. However, existing solutions, such as software-based seat reservations and sensors-based occupancy detection, have proven to be inadequate in effectively addressing this problem. In this study, we propose a novel approach: a serial dual-channel object detection model based on Faster RCNN. This model is designed to discern all instances of occupied seats within the library and continuously update real-time information regarding seat occupancy status. To train the neural network, a distinctive dataset is utilized, which blends virtual images generated using Unreal Engine 5 (UE5) with real-world images. Notably, our test results underscore the remarkable performance uplift attained through the application of self-generated virtual datasets in training Convolutional Neural Networks (CNNs), particularly within specialized scenarios. Furthermore, this study introduces a pioneering detection model that seamlessly amalgamates the Faster R-CNN-based object detection framework with a transfer learning-based object classification algorithm. This amalgamation not only significantly curtails the computational resources and time investments needed for neural network training but also considerably heightens the efficiency of single-frame detection rates. Additionally, a user-friendly web interface and a mobile application have been meticulously developed, constituting a computer vision-driven platform for detecting seat occupancy within library premises. This research effectively addresses the persisting issue of seat occupancy management within library systems. Noteworthy is the substantial enhancement in seat occupancy recognition accuracy, coupled with a reduction in computational resources required for neural network training, collectively contributing to a considerable amplification in the overall efficiency of library seat management.

## I. INTRODUCTION

As a public place, the library is a place where students and teachers learn, read and consult literature. Library seat occupancy is a phenomenon that a person puts personal belongings on the seat to indicate that the seat is occupied, but the owner is not on the seat. This phenomenon will make the use of library seat resources more tense, and aggravate the contradiction of library resource shortage. Especially in the exam season, this phenomenon is more common. Traditional librarians use patrol and warning methods to solve this problem, but this will undoubtedly increase the workload of staff. Therefore, how to use modern technology to solve the problem of library occupancy is a challenge.

Currently, there are two commonly used seating solutions. The first scheme is that the students can use mobile software to reserve seats in advance, and those who have not made an appointment cannot use seats. This scheme is widely used in colleges and universities [1]-[4]. However, the survey shows that most of the staff use their seats without making an appointment, which obviously does not solve the problem of seat occupancy in the library. The second scheme is to detect the people on the seat with various sensors. Tunahan et al. (2022) [5] invented a seat monitoring system that detects people's actions through Passive Infrared Sensor (PIR) and then judges the seat status. Daniel et al. (2019) [6] invented a detection system, which detects seat pressure through a pressure sensor, conducts identity recognition through an RFID sensor, and then detects seat status. But installing a sensor system for each seat means high costs. Moreover, the robustness of the sensor is poor, and it is easy to be disturbed or even damaged by other external factors, which requires frequent inspection and maintenance. Aiming at the drawbacks of the above schemes, this paper innovatively proposes a serial dual-channel detection model based on Faster RCNN algorithm for serial object detection and object classification. The system has the advantages of low cost, high detection accuracy and good visualization effect.

Object detection is one of the most important problems in the field of computer vision. Traditional object detection algorithms need to set features manually and the process is tedious. Besides it used the shallow classifier which has low accuracy in complex environments. The development and application of deep learning has greatly improved the performance of object detection. At present, the most advanced object detection algorithms are YOLO and Faster RCNN. As shown in TABLE I, in recent years, the two algorithms have been widely used in unmanned driving, agriculture, medical and other fields. Yang et al. (2022) [7] used Faster RCNN algorithm to detect traffic vehicles in order to reduce traffic accidents. Panda et al. (2021) [8] used Faster RCNN algorithm to detect peanut leaf disease, greatly improving the yield of peanuts. Khasawneh et al. (2022) [9] used the Faster RCNN algorithm to achieve the detection of EEG, which contributes to the prevention and treatment of cancer. Jiang et al. (2022) [10] used YOLOv5 algorithm to detect objects in infrared images taken by UAVs, which greatly improved the search performance of UAVs. Chen et al. (2021) [11] used YOLOv5 algorithm in face recognition to improve the accuracy of face recognition. Liang et al. (2022) [12] deployed YOLOv3 algorithm on NVIDIA Jetson drive platform and proposed a new solution for obstacle avoidance of unmanned driving. YOLO algorithm is a one-stage algorithm with extremely fast detection speed, which is suitable for real-time dynamic detection. Faster RCNN algorithm is a two-stage algorithm, whose detection speed is slower than YOLO [13], but its network complexity makes its accuracy higher than YOLO, which is suitable for single frame image detection. In general, Faster RCNN is more accurate than YOLO in detecting small objects due to its ability to detect more candidate frames and handle the size and position variations of objects. On the other hand, YOLO may be more accurate than Faster R-CNN in detecting large objects because it considers the whole image in the prediction process instead of processing each region individually. This enables YOLO to capture the contextual information of large objects more effectively. Therefore, Faster RCNN algorithm is more suitable for this model. The COCO dataset built by Lin et al. (2014) [14] has greatly reduced the time for researchers to tag photos, making migration learning in complex scenes faster and more accurate. But at present, no scholar has applied target detection to library occupancy detection. Therefore, this paper uses the deep learning method to carry out research on the detection of library occupancy.

TABLE I. Comparison of two target detection algorithms

| Algorithm | Type | Advantage | Application |
|---|---|---|---|
| Faster RCNN | double-stage algorithm | Higher precision | Traffic Detection [7], Agriculture [8], Medical Science [9] |
| YOLO | One-stage algorithm | Faster | UAV Search [10], Face Detection [11], Autonomous Cars [12] |

However, deep learning network training requires a large number of training sets, especially for a complex scene. It is difficult to obtain library scenes. Virtual reality technology can build various scenes and personalized characters through computers, and can provide massive training sets for deep network models. After decades of development, virtual reality technology has become highly reductive. Liu. (2022) [15] used virtual reality technology to build a virtual fitting room and virtual characters, making it more convenient to buy clothes online. Similarly, virtual reality technology is also widely used in the medical field. The use of virtual characters for remote surgery and diagnosis reduces the time cost of treatment. The construction of human organs and the simulation of physical activities are of great significance to medical research [16-18].Virtual reality technology also makes it possible to teach in a virtual environment that is unimaginable in the physical classroom, such as entering virtual laboratories, virtual classrooms and virtual conference rooms [19,20].The huge possibility of accessible virtual technology will make it possible to break the boundaries of physical reality.、

On the basis of the above research, as shown in Fig. 1, this paper proposes a du-al channel detection model based on Faster RCNN algorithm for serial target detection and object recognition. And integrate image acquisition, image detection and seat management platform to develop a library seat intelligent management platform. The system consists of image input terminal, dual channel detection model and user terminal. After image segmentation and pre-processing, each sub image first performs target detection, and no sub image of human is detected for object recognition. If a book is recognized, it is determined that the seat is suspected of occupying a seat. The innovation points of this paper are summarized as follows:

（1）A dual channel detection model is proposed, in which target detection and object recognition are carried out serially to reduce the labeling of data sets. Only people and books are trained online, which greatly reduces the cost of computing re-sources and time.

（2）Use the Unreal Engine 5 to build virtual scenes and create characters. You can generate data sets of various scenes, various characters and various perspectives at will.

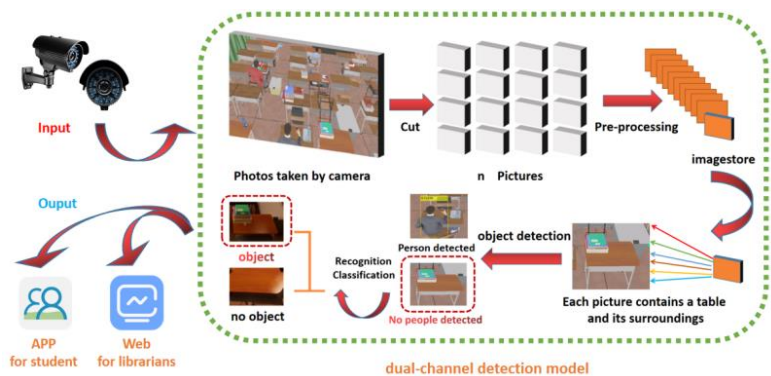

Fig. 1. Flow chart of library intelligent seat management system

The rest of this article is organized as follows: Section 2 shows how the virtual engine builds the dataset. Section 3 shows the selection and training of network in object detection. Section 4 shows the selection and training of network in object classification. Section 5 shows the Web interface and APP of the system, and conducts a live test of the system. Section 6 summarizes the full text and puts forward prospects.

II. VIRTUAL ENGINE DATA SET CONSTRUCTION

*A. Virtual scene construction*

The process of creating a virtual scene using Unreal Engine 5 and open source materials can be broken down into several steps. Firstly, all the open source materials required are collected, including classrooms, seats, chairs, books, boxes, schoolbags, and people sitting and standing. Then, all the collected materials are imported into the same project. Using the rendering feature of UE5, the classroom, table, and chair are added to the viewport and their relative positions and size parameters are adjusted. To enhance the realism of the virtual scene, other materials are continued to be added to the viewport, such as white plastic boxes and curtains. Afterwards, the material of different items is standardized using the material editor. Finally, the brightness of the scene is adjusted and the construction of the virtual scene is completed.

* Corresponding authors:
E-mail address: myang@xidian.edu.cn

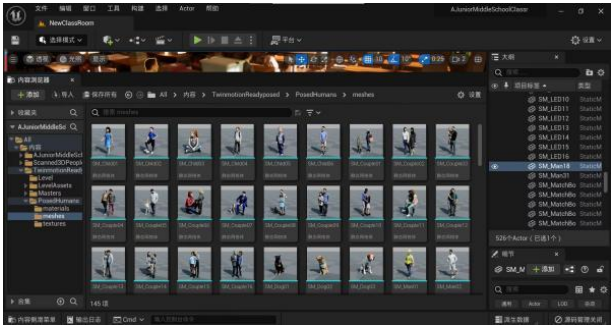

(a) Material collection of people with different sitting positions

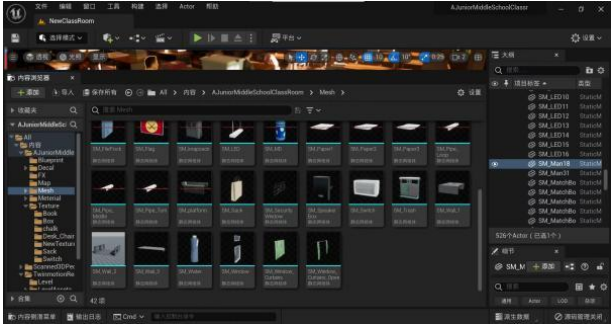

(b) Material collection of different items

Fig. 2. UE5 Interface

### B. Data collection under different viewing angles and illumination

In this section, we simulate the phenomenon of occupied and non-occupied seats in a library by adding different materials and adjusting different viewpoints and lighting to obtain a virtual dataset. The simulation of the non-occupancy phenomenon is done as follows: randomly select different items from the item network diagram to add to the viewport and drag them to the desk. Then randomly select a character from the sit-stand character material set and add it to the seat. The simulation of seat occupation is as follows: randomly select items from the item material set and add them to different desks without other processing. Finally, simulate the viewpoint detected by the camera in a real library scene. Adjust the camera position to the top view and fix it. Repeat the above steps, and change the camera position to change the view angle, change the light intensity, and finally a large virtual reality dataset can be obtained.

Fig. 3 shows some examples of the virtual dataset we have constructed. Fig. 3(a) and Fig. 3(b) have fixed camera views at the top of the back of the classroom, and the pictures contain 5 and 3 people respectively. The fixed camera view in Fig. 3(c) Fig. 3(d) is located right above the classroom and contains eight desks. In Fig. 3(c), four seats are occupied, and in Fig. 3(d), two seats are occupied. Adjust the light intensity in Fig. 3(e) and Fig. 3(f), where 4 seats are occupied in Fig. 3(e), and 5 seats are occupied in Fig. 3(f).

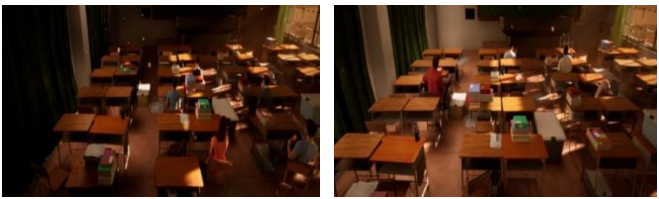

(a) 3 people in classroom    (b) 5 people in classroom

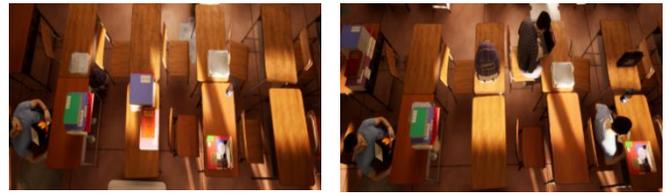

(c) Weak light, 4 seat occupancy phenomena    (d) Weak light, 2 seat occupancy phenomena

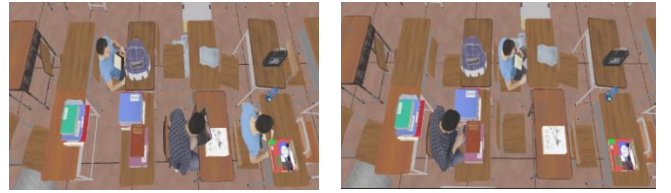

(e) Strong light, 4 seat occupancy phenomena    (f) Strong light, 5 seat occupancy phenomena

Fig. 3. These images are some examples of the virtual datasets we have constructed

### C. Data pre-processing

We collected a total of 366 images, including 266 images actually taken and 100 constructed based on virtual reality. The actual shooting location is the library of Xidian University (Xi'an, Shaanxi, China), Shandong University (Jinan, Shandong, China) and South China University of Technology (Guangzhou, Guangdong, China). In order to further expand the dataset and enhance the robustness of the model in processing different data, image pre-processing is required for the dataset.

In actual library image acquisition, the camera image acquisition process can easily degrade the image quality due to some uncontrollable factors. One of the most influential factors for image recognition is illumination. When the viewing angle is fixed, three different lighting conditions including morning, evening and night can have an impact on person recognition. In camera imaging, mis-exposed photos can be divided into two categories as follows:

(1) When the light intensity is too low, the overall brightness of the captured image is low due to underexposure, and the detailed information in the dark areas of the image cannot be clearly presented.

(2) When the light intensity is too high, the overall brightness of the captured image is high due to overexposure, and the detail information of the bright area of the image cannot be clearly presented.

Since the identification of library seats does not require too fine a brightness adjustment for local areas, this section selects a histogram equalization-based method for overall image pre-processing. Histogram equalization is the most basic and stable image processing method, which is also the most simple, intuitive and effective. The main purpose of this method is to redistribute the gray values of pixels with different luminance in a low-light image. By expanding the range of grayscale values in the image, the pixel values are evenly distributed in the image. This improves the brightness and contrast of the original input image, reveals the details hidden in the dark areas of the image, and effectively improves the visual effect of the image.

For color images, histogram equalization can be performed for R, G, and B components respectively, but this may lead to color distortion of the resulting image. Therefore, this paper converts RGB space into HSV (hue, saturation and brightness), and equalizes the histogram of V component to ensure that the image color is not distorted. This part calls the Matlab function histeq(), but uses the self-coding method for processing. The specific effect is shown in Fig. 4.

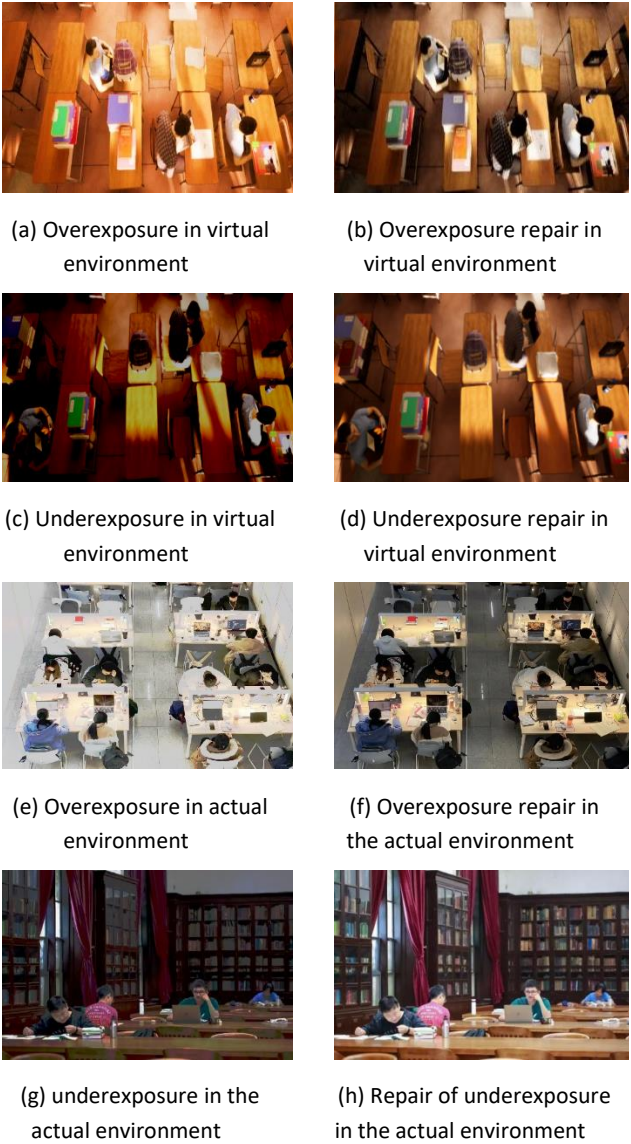

(a) Overexposure in virtual environment
(b) Overexposure repair in virtual environment
(c) Underexposure in virtual environment
(d) Underexposure repair in virtual environment
(e) Overexposure in actual environment
(f) Overexposure repair in the actual environment
(g) underexposure in the actual environment
(h) Repair of underexposure in the actual environment

Fig. 4. Comparison before and after image preprocessing

Through the repair of overexposure and underexposure, the recognition rate of people and objects can be effectively distinguished. Fig. 5(a) shows that people and books cannot be accurately identified due to insufficient exposure, and the accuracy rate of identifying people is only 75%. However, Fig. 5(b) shows that 100% recognition can be achieved after correction.

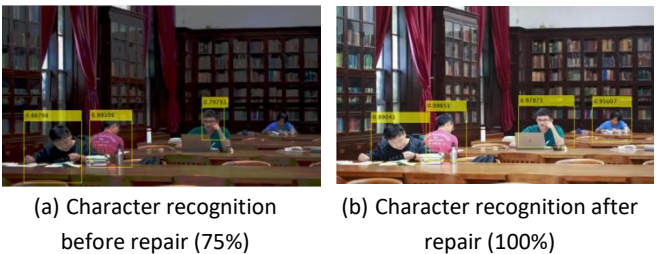

(a) Character recognition before repair (75%)
(b) Character recognition after repair (100%)

Fig. 5. Character recognition before and after repair

Therefore, in this paper, before the person detection, the histogram equalization of the image captured by the camera is carried out, and the image brightness is automatically adjusted, and then the part of occupancy detection is entered.

### III. OBJECT DETECTION BASED ON FASTER RCNN

#### A. Faster R-CNN core architecture

Faster R-CNN is an object detection algorithm based on classification, which belongs to monocular two-stage detection algorithm. It inherits the idea of traditional object detection and treats object detection as a classification problem. Faster RCNN goes further on the basis of Fast RCNN, and integrates the candidate box generation part into the CNN network, so that the four parts of candidate box generation, feature extraction, candidate box classification, and candidate box boundary regression are all combined in a CNN network. This avoids step by step training and truly realizes end-to-end object detection. The Region Proposal Network (RPN) is utilized to generate candidate bounding boxes, while the ROI Pooling layer maps these generated boxes onto a fixed-size feature map and extracts features. The predicted coordinates and category scores of the bounding boxes are ultimately produced by two fully connected layers. The main network structure of Faster RCNN is shown in Fig. 6, including the following four parts: feature extraction network, region generation network, ROI pooling layer, and classification regression layer.

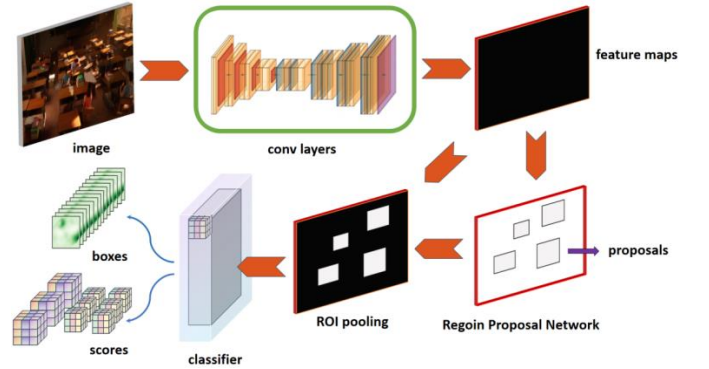

Fig. 6 Faster R-CNN Network Architecture

#### B. Standard accuracy and loss value of evaluation index

Single object detection belongs to the binary classification problem. The result which is detected is called Positive, the undetected is called Negative. The correct pre-diction of the classifier is marked as True, and the wrong prediction is marked as False. These four basic terms are combined to form four basic elements of evaluation object detection, as shown in TABLE II.

TABLE II. COMPARISON OF TWO TARGET DETECTION ALGORITHMS

| Real | Forecast results | |
|---|---|---|
| | Positive | Negative |
| Positive | TP (Ture Postive) | FN (Fasle Neagtive) |
| Negative | FP (False Postive) | TN (Ture Neagtive) |

TP and TN are both accurate predictions. We call their proportion in all prediction results Accuracy. The formula is:

$$\text{Accuracy} = \frac{TP + TN}{TP + TN + FP + FN} \quad (1)$$

L1 Loss is also called Mean Absolute Error, or MAE, which measures the average error amplitude of the distance between the

predicted value and the true value. The formula is:

$$Loss = \frac{1}{n}\sum_{i=1}^{n}|y_i - f(x_i)|  \quad (2)$$

where n is the number of samples, $y_i$ is real border data, $f(x_i)$ is forecast box data.

The training results in this paper use accuracy and loss as evaluation indicators.

### C. Pre-selection of characteristic network

For Faster RCNN algorithm, the selection of feature extraction network directly affects the performance of the network. Through a series of convolution and pooling operations, the feature extraction network can extract features from input images and generate feature maps. More and more convolutional neural networks (CNN) with superior performance, such as AlexNet [21], GooLeNet [22], SqueezeNet [23], VGGNet [24] and ResNet [25], have been proposed by scholars.

This section uses AlexNet, GooLeNet, SqueezeNet, VGG19, VGG16 and ResNet18 as the pre-training network. The training set includes different scenes, such as individual portrait Fig. 7(a) [26], field scene Fig. 7(b) and library scene Fig. 7(c), to ensure the diversity of dataset. Through pre-training, the network structure with the best performance and the strongest generalization ability is evaluated.

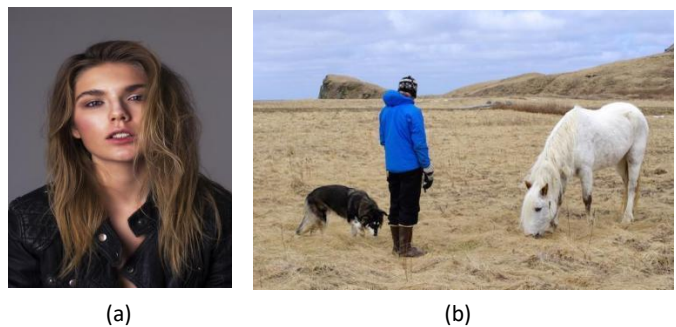

Fig. 7. Examples of training set

The accuracy and loss are used as evaluation indicators to pre-train the six net-works respectively. A training set containing 88 images and a test set containing 22 images are used in this training. The training set and test set were assigned in a ratio of 8:2. The training times of the neural network are 20, and a total of 1760 iterations. The whole training process is carried out in the Matlab 2022 (b), using the single GPU mode, and the training results are shown in Fig. 8 and Fig. 9.

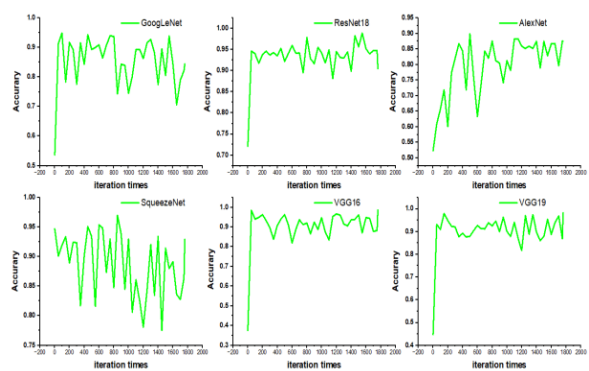

Fig. 8. Accuracy of six kinds of network training

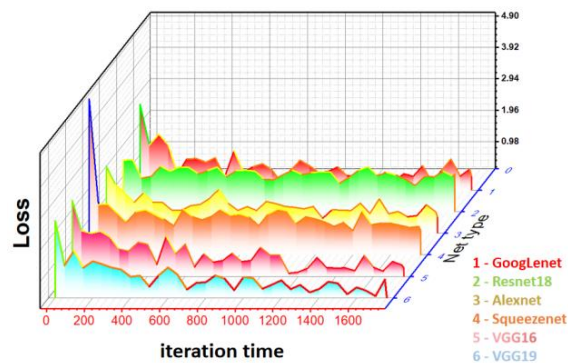

Fig. 9. Six network training loss

Fig. 8 shows the changes in the accuracy of the six networks during the training iteration. The accuracy changes curves of ResNet18 and VGG19 networks are relatively stable, indicating that the convergence speed of both networks is fast. Fig. 9 shows the changes of the loss of the six networks during the training iteration. VGG16 and VGG19 networks decline fastest, indicating that they converge fastest.

The maximum accuracy, average accuracy, minimum loss and average loss in the six network iteration processes are calculated. The results are shown in TABLE III and TABLE IV. TABLE III shows the maximum and average accuracy rates in different network training processes. The maximum accuracy that VGG19 can achieve is the maximum, and that of AlexNet is the minimum. This is because VGG19 has more network layers and uses smaller convolutional kernels to better preserve the image characteristics. On the whole, the average accuracy of ResNet18 is the highest, while that of AlexNet is still the lowest. This is because ResNet18 uses a deeper network and has a deeper classification accuracy. TABLE IV shows the minimum and average loss in different network training processes. On the whole, the average loss of VGG19 is the smallest and that of SqueezeNet is the largest. This may be because there are many full connection layers in VGG19 network, which makes the output score at the classification stage higher and the fitting degree with the real samples higher.

TABLE III. MAXIMUM ACCURACY AND AVERAGE ACCURACY OF SIX KINDS OF NETWORK

| Net type | Accuracy max | Accuracy average |
|---|---|---|
| GoogLeNet | 94.91% | 85.37% |
| ResNet18 | 98.91% | 93.31% |
| AlexNet | 88.28% | 79.69% |
| SqueezeNet | 94.83% | 88.69% |
| VGG16 | 98.48% | 90.45% |
| VGG19 | 98.85% | 92.28% |

TABLE IV. MINIMUM AND AVERAGE LOSSES OF SIX KINDS OF NETWORK TRAINING

| Net type | Loss min | Loss average |
|---|---|---|
| GoogLeNet | 0.3770 | 0.8895 |
| ResNet18 | 0.5851 | 0.8320 |
| AlexNet | 0.5990 | 1.0189 |
| SqueezeNet | 0.8914 | 1.3795 |
| VGG16 | 0.2323 | 0.8089 |
| VGG19 | 0.1343 | 0.7472 |

According to the comprehensive evaluation of accuracy, loss and convergence speed, VGG19 and ResNet18 networks have the best performance. According to the actual detection capability of the network, use the trained network to detect and com-pare the test set, and the results are shown in Fig. 10. Fig. 10(a) and Fig. 10(b) show that under the condition of single person detection, the confidence of VGG19 recognition is far greater than ResNet. Fig. 10(c) and Fig. 10 (d) show that the recognition rate of ResNet is lower than VGG19 and the recognition confidence of the same person is still lower than VGG19 when the number of people to be detected is in-creased. The dataset in Fig. 10(e) and Fig. 10(f) continue to increase the number of people to be detected, at this time, the clarity of everyone in the picture is greatly reduced. There are 22 people in the picture, 11 people are detected in Fig. 10(e), the recognition rate is 50%, and 17 people are detected in Fig. 10(f), the recognition rate is 77%. To sum up, the actual detection performance of VGG is better than ResNet network, so VGG is selected as the feature extraction network.

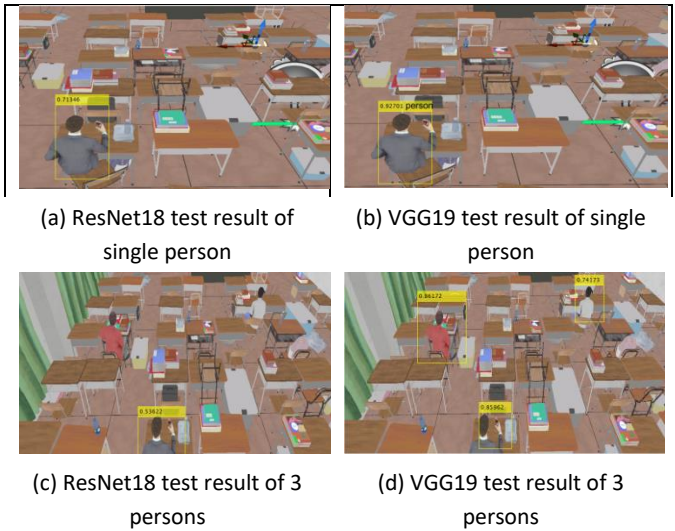

(a) ResNet18 test result of single person

(b) VGG19 test result of single person

(c) ResNet18 test result of 3 persons

(d) VGG19 test result of 3 persons

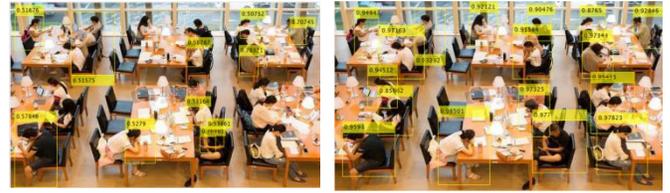

(e) ResNet18 test result of multiple persons

(f) VGG19 test result of multiple persons

Fig. 10. Test results of ResNet 18 and VGG19

### D. Network training

Section 3.3 uses a small sample size to select a suitable convolutional neural network. This section aims to improve the adaptability of the network to complex scenarios such as libraries, and to verify that the network trained using a training set containing both virtual and real dataset outperforms the network trained using only real dataset as the training set. The number of dataset is increased, especially the photos of students studying in the library. There are two datasets in this time, a and b. The dataset a consists of 103 virtual reality constructed images and 103 realistically captured images. The dataset b consists of 206 realistically captured images, of which 103 images are duplicated with the images in the dataset a. Dataset a and b both have a training set to test set allocation ratio of 8:2. The whole training process is carried out in Matlab 2022 (b), using the single GPU mode. Fig. 11(a) and Fig. 11(c) show the change of accuracy with the number of iterations in the training process. Since migration learning is conducted on the basis of pre-trained networks, the original network dataset is COCO set [13], so the accuracy of the network is extremely high. However, there are many jitters in the curve, which indicates that there is a problem of local optimization in this training. Fig. 11(b) and Fig. 11(d) show the change of the loss with the number of iterations in the training process. The network trained using dataset a starts to converge at about 2000 times, and the convergence speed is slow. However, the network trained using dataset b converges in a few dozen iterations, and convergence is extremely fast. Fig. s 11(e) and 11(f) display the PR and mAP curves, respectively, of the training process using different training sets. These curves demonstrate that the network trained with both virtual and real dataset outperforms the network trained solely with real dataset. TABLE V calculates the highest accuracy, average accuracy, minimum loss and average loss of training. Compared with the training results of small and medium-sized samples in the previous section, with the increase of dataset, the network performance is getting better. In terms of the training accuracy, the accuracy of the networks trained with dataset a and dataset b are both above 91%, which is almost the same. However, in terms of training loss, the average loss of the network trained with dataset a is only 0.6165, but the average loss of the network trained with dataset b is as high as 1.2659. This shows that the performance of the network can be improved by training the network with both virtual and real datasets.

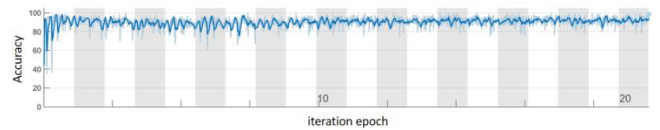

(a) Training accuracy curve based on virtual dataset and realistic training sets

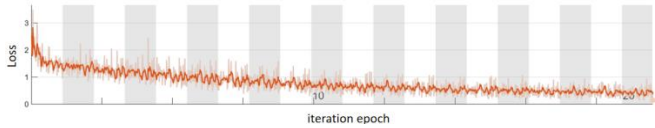

(b) Training loss curve based on virtual dataset and realistic training sets

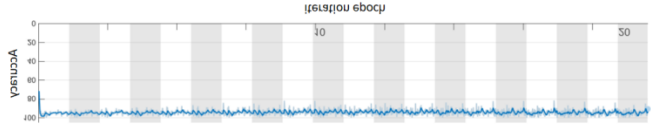

(c) Training accuracy curve based on realistic training set only

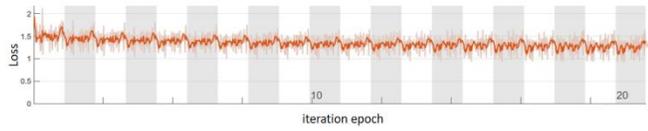

(d) Training loss curve based on realistic training set only

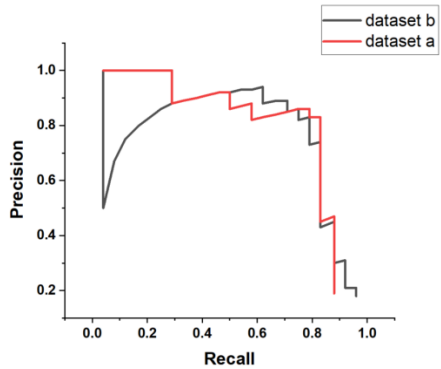

(e) Training PR curve based on different type of training set

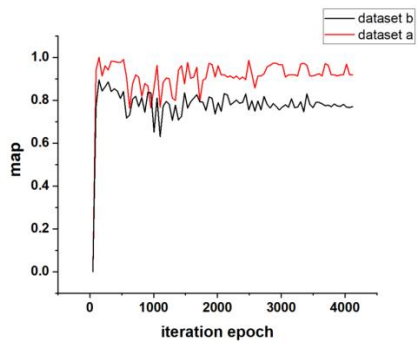

(f) Training mAP curve based on different type of training set

Fig. 11. VGG19 Training Process Curve based on different type of training set

TABLE V. TRAINING ACCURACY AND LOSS

| Training set type | Accuracy max | Accuracy average | Loss min | Loss average |
|---|---|---|---|---|
| virtual dataset and realistic training sets | 98.36% | 91.85% | 0.1165 | 0.6165 |
| realistic training set only | 97.94% | 94.56% | 1.2036 | 1.2659 |

*E. Network test*

This section uses the enhanced network to test the test set. The results are shown in Fig. 12. Fig. 12(a) shows that when the test object is a single person, the recognition confidence is 99%. Fig. 12(b) increases the number of people tested to 4, the number of people identified is 4, and the recognition rate is 100%. Fig. 12(c) continues to increase the number of people tested to 6, the number of people identified is 6, and the recognition rate is 100%. Fig. 12(d) continues to increase the number of people tested to 18, and the number of people identified is 18, with a recognition rate of 100%. From the above results, the recognition rate of this network can reach 100%. Fig. 12(e)-12(h) simulates the impact of different perspectives and lighting conditions on network testing. The results show that the viewing angle and lighting conditions will not affect the performance of the network.

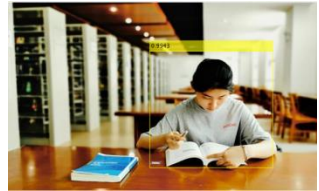
(a) Test result of single person

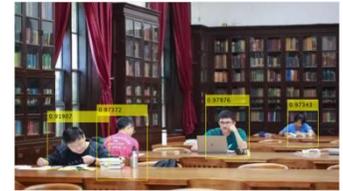
(b) Test result of 4 persons

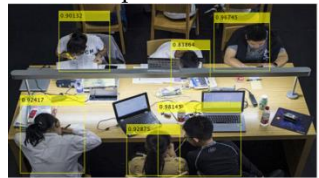
(c) Test result of 6 persons

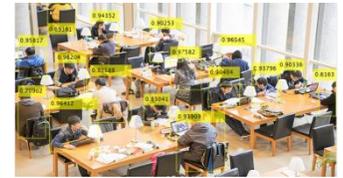
(d) Test result of multi persons

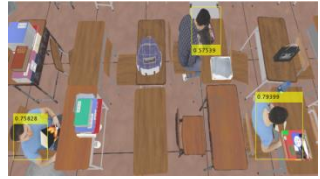
(e) Simulated top view, morning (strong light) test result

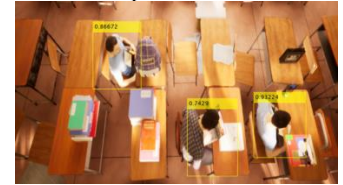
(f) Simulated top view and afternoon (weak light) test result

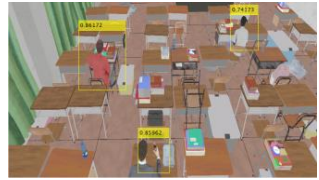
(g) Simulated side view, morning (strong light) test result

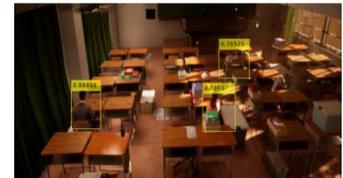
(h) Simulated side view, afternoon (low light) test result

Fig. 12. Network Test Results

## IV. OBJECT CLASSIFICATION

Based on the person image information that has been identified in Section 3, the acquired images are firstly aliquoted based on the distribution of study room tables and chairs. Based on the result of object detection, the pixel blocks of images without people next to the tables and chairs in the photos are filtered. And classify them into two categories, objects and no-objects, to conclude whether the seat is suspected to be occupied or not.

## A. Image selection and processing

The training data are stored in certain subfolders according to the number of categories, and the name of each subfolder is the label of the corresponding category: objects and no-objects. The selected dataset has better accuracy when the selected dataset has 100 images per category and 200 images in total. The dataset we selected in the item classification are both collected from the Internet and include dataset built based on section 2. In order to ensure that the dataset has high application value and practical significance for reality, this paper still focuses on the real scene data in the construction of the migration learning dataset, with the virtual reality constructed images as the auxiliary. The ratio of image data from the two sources is 7:3 for the following experimental design and extension application.

## B. Selection of neural network structure and parameter setting

Looking at the effect of classical deep networks and network layer construction, in order to ensure the suitability of network applications, this paper selects a total of five networks, AlexNet, GoogLeNet, VGG16, ResNet50 and SqueezeNet, for testing and analysis based on a comprehensive consideration of both accuracy and execution efficiency. Finally, the network with the highest fitness and accuracy is selected from these five networks as the main framework architecture for our model. The classical network comparison diagram is shown in Fig. 13.

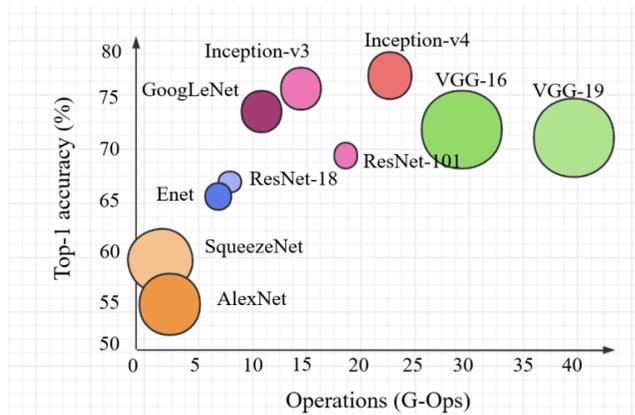

Fig. 13. Classical network performance comparison

In the initial network selection and testing, the five selected networks AlexNet, GoogLeNet, VGG16, ResNet50 and SqueezeNet are trained and evaluated in this section. To address the differences in their network recognition results, this section obtains the training time and predicted classification accuracy of the five networks under different parameter conditions by fixing a series of parameter values, so as to deter-mine which network is the best network for object classification. The training options are pre-trained so as to select the most stable values for most of the networks' recognition rates, and are set well before training the model. The parameters and their values are shown in TABLE VI.

TABLE VI. TRAINING PARAMETERS

| Optimizing Algorithm | Stochastic Gradient Descent with Momentum |
|---|---|
| Mini Batch Size | 32 |
| Maximum Epochs | 40 |
| Learning rate | 0.0005 |
| Classification rate | 7/3 |

These parameter settings were trained with six different networks. Only the image set classification ratio was considered as the basis for comparing the network performance, and the specific data is shown in TABLE VII.

TABLE VII. Comparison of data recognition under different networks

| Network Category | Training Time | Accuracy of mini-Batch | Loss of mini-Batch | Successful Rate |
|---|---|---|---|---|
| AlexNet | 1: 29 | 100% | 0.0018 | 0.936 |
| GoogLeNet | 1: 54 | 98.76% | 0.0043 | 0.925 |
| VGG16 | 2: 36 | 100% | 0.0006 | 0.930 |
| ResNet50 | 1: 57 | 96.22% | 0.0341 | 0.824 |
| SqueezeNet | 1: 26 | 96.45% | 0.0018 | 0.833 |

The experimental data revealed that the pre-trained networks AlexNet, GoogLeNet and VGG16 provided better stability and higher recognition rate during RCNN training. Although VGG16 has higher recognition than GoogLeNet, its excessive training time and high computing cost are less suitable for small-scale library occupancy detection, so its application is firstly excluded.

Compared with AlexNet, SqueezeNet has a shorter overall training time, but its stability in object classification is inferior to AlexNet. Although both networks have an accuracy of over 90% in recognition accuracy, the highest recognition accuracy is still achieved by AlexNet, at 93.6%. Although the accuracy of the GoogLeNet network is also high, it is prone to overfitting due to its significantly longer training time than AlexNet and the limitations of the training set.

Specifically, AlexNet has been shown to be one of the simplest networks that can be trained in detection applications and perform with high accuracy. From the perspective of library occupancy detection system application, its most valuable factors and benchmarks for recognition judgment are recognition accuracy and network stability, so AlexNet is chosen as the network structure in this section.

As shown in Fig. 14, AlexNet is a pre-trained eight-layer CNN architecture consisting of five convolutional layers and three fully connected layers. A nonlinear and nonsaturated ReLU function is selected for the activation function, and local response normalization (LRN) is used to suppress other neurons with smaller feedback, which enhances the generalization ability of the model. The AlexNet uses overlapping pooling, which makes each pooling have overlapping parts and can avoid the overfit-ting phenomenon.

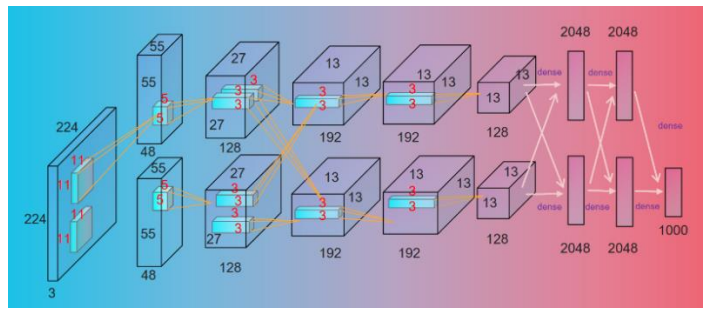

Fig. 14. AlexNet Network Structure diagram

## C. Model training

Since the parameter initial learning rate has a large impact on the accuracy of the training model, its value is too large to cause the model to not converge, and too small to cause the model to converge particularly slowly or fail to learn. After determining the initial learning rate boundary value, three values of 0.0001, 0.0005 and 0.001 are selected as the base learning rate in this paper. Since the multi-GPU collaborative feature of AlexNet can greatly improve the training speed and use dropout to effectively reduce overfitting, the number of training iterations is increased appropriately to avoid overfitting and maximize the accuracy of object classification. Finally, 40 is chosen as the maximum number of iterations. TABLE VIII shows the training results.

By increasing the number of iterations or changing the ratio of training and testing images and the base learning rate, different evaluation criteria such as runtime, mini-batch accuracy, mini-batch loss, and final accuracy can be obtained for the dataset under different conditions.

By varying the base learning rate, the maximum number of iterations, and the training/prediction classification ratio, this section yields data on the training time, mini-batch accuracy, mini-batch loss, and final accuracy required for image classification under different combinations. TABLE VIII shows that as the classification ratio de-creases, the required training time gradually becomes larger, while the accuracy rate shows different trends at different learning rates. Taking the base learning rate of 0.001 as an example, its recognition rate is gradually lower as the classification ratio de-creases, indicating the occurrence of overfitting; however, when the base learning rate is 0.0001, it shows the opposite trend, indicating that it is still not saturated.

The final classification accuracy is highest when the base learning rate is 0.001,the classification ratio is 6/4, the base learning rate is 0.0005 and the classification ratio is 7/3, the final classification accuracy is 0.936. However, the former parameter setting requires less training time and has a lower mini-batch loss, making it the optimal setting for image classification.

TABLE VIII. TRAINING ACCURACY WITH DIFFERENT PARAMETERS

| Learning Rate | Classification ratio | Training time(min) | Accuracy of mini-Batch | Loss Rate of mini-Batch | Successful Rate |
|---|---|---|---|---|---|
| 0.0010 | 6/4 | 0:35 | 100% | 0.0012 | 0.936 |
| 0.0010 | 7/3 | 1:17 | 100% | 0.0057 | 0.900 |
| 0.0010 | 8/2 | 1:42 | 98.76% | 0.0010 | 0.813 |
| 0.0010 | 9/1 | 2:07 | 97.89% | 0.0054 | 0.867 |
| 0.0005 | 6/4 | 0:42 | 100% | 0.0006 | 0.900 |
| 0.0005 | 7/3 | 1:29 | 100% | 0.0018 | 0.936 |
| 0.0005 | 8/2 | 1:50 | 99.12% | 0.0025 | 0.900 |
| 0.0005 | 9/1 | 2:25 | 100% | 0.0045 | 0.875 |
| 0.0001 | 6/4 | 1:15 | 98.63% | 0.1206 | 0.833 |
| 0.0001 | 7/3 | 1:51 | 100% | 0.0319 | 0.850 |
| 0.0001 | 8/2 | 2:24 | 97.56% | 0.0682 | 0.875 |
| 0.0001 | 9/1 | 3:01 | 100% | 0.0781 | 0.799 |

To visually determine the accuracy of object classification, the concept of confusion matrix is introduced here. The confusion matrix is a situation analysis table for summarizing the prediction results of classification models in deep learning. In library occupancy detection, the confusion matrix is divided into four quadrants in order of top to bottom and left to right for exploration. As shown in Fig. 15, the number of correct predictions is represented in the blue quadrant: the first quadrant is the number of objects correctly identified, and the fourth quadrant is the number of objects correctly identified without objects. The second and third quadrants represent mis-classification: the second quadrant element is the number of incorrectly identified objects as no objects, and the third quadrant element is the opposite. It can clearly reflect the accuracy of image classification and the problems that occur. When the number of objects in the blue quadrant is larger, it means that the model has a higher recognition rate under this condition. For example, with a learning rate of 0.001 and a classification ratio of 6/4, the number of objects accurately identified is 52 (26+26), but the number of objects incorrectly identified as having objects on the table is 5, and the number of objects incorrectly identified as no objects is 6. Fig. 15 shows the confusion matrix for each of the 12 parameter conditions.

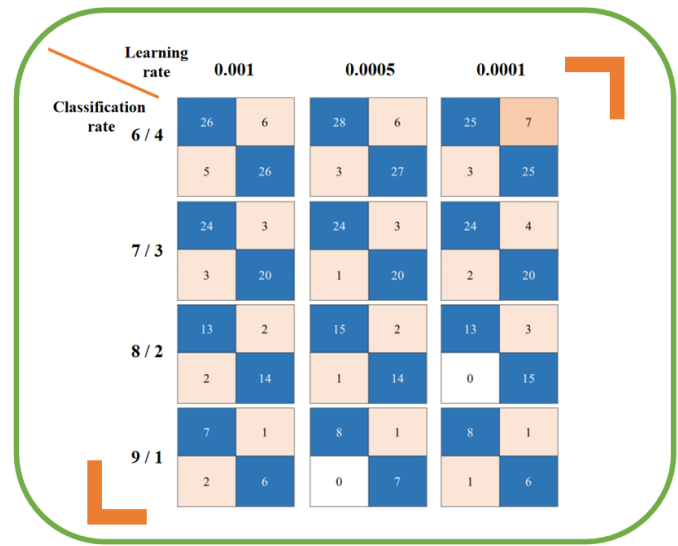

Fig. 15. Confusion matrix of different parameters

After continuous testing of the dataset, the success rate was 0.936 after selecting the optimal parameters to set the conditions. The trained network was used to classify the images and the results are shown in Fig. 16. From the test classification results, the trained network has a more satisfactory classification rate for the presence or absence of objects, and the confidence level is high enough to undertake the task of checking the classification of the presence or absence of objects on the seats, so that it can well determine whether the students are suspected of occupying the seats.

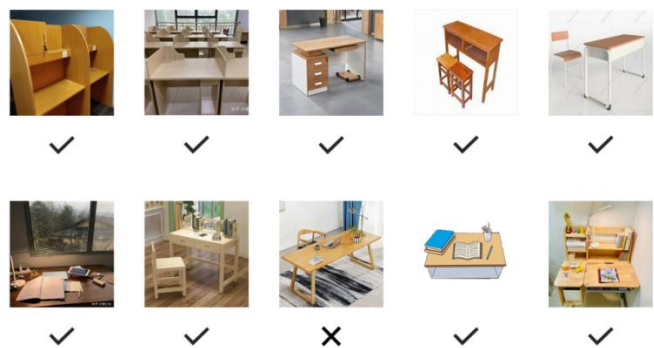

Fig. 16. Test results of object classification

## V. EXPERIMENT AND RESULT DISPLAY

### A. System Hardware Module Construction

In order to verify the validity of the model, the model is tested in this section. As shown in Fig. 17, the camera model used is Logitech C270 HD Webcam [27]. The maximum resolution is 720 p / 30 fps, and photos can be transferred to a computer via WiFi, Bluetooth, etc. The test platform is Matlab 2022 (b) and the test site is the study room on the 2nd floor of the library of Xidian University (Xi'an, Shaanxi, China).

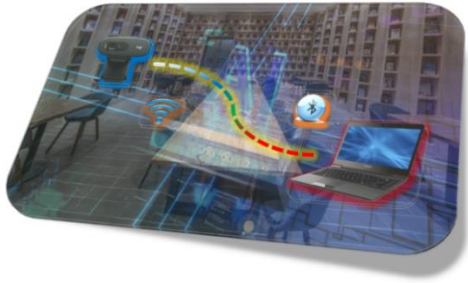

Fig. 17. System hardware block diagram

### B. System Software Module

For the presentation of the system output, a Web interface for librarians and an APP for students are designed. The system achieves end-to-end functionality.

Fig. 18 shows the Web interface designed using Matlab APP Designer, which can be used by the librarian. The images can be imported from the camera and saved to the database of the computer, or any image can be imported manually. The image box on the left shows the imported photos, and the image box on the right shows the segmented sub-images. √ means there is no occupancy at the seat, × means there is occupancy at the seat. When the button is green, it means the detection is completed, and it turns red means the detection is in progress.

Fig. 19 shows the app designed for students, written in JavaScript and running on Android 6.0.1 and above. Fig. 19(b) indicates that seats with occupancy will be shown in red, blue means that the seat is not occupied, and gray means that the seat is not used. If the student is occupying the seat, the APP will make a message alert via cell phone. In addition, the APP also adds the announcement function, which can re-lease some school-related information such as recruitment and notification.

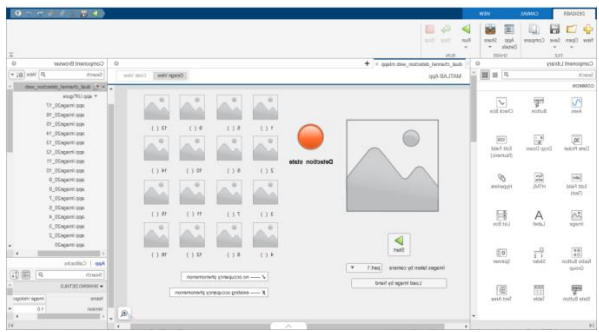

(a) Web interface design

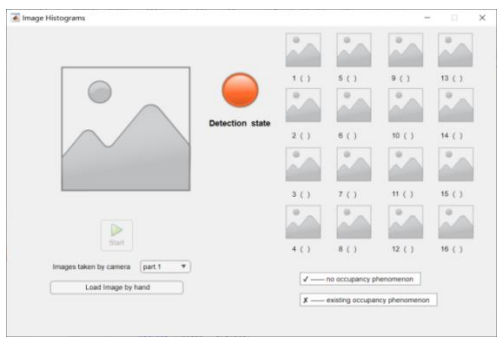

(b) Web interface

Fig. 18. Web interface

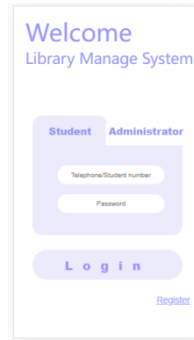 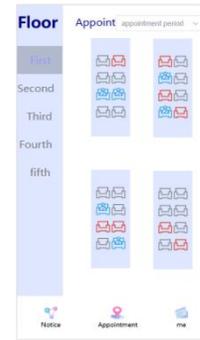 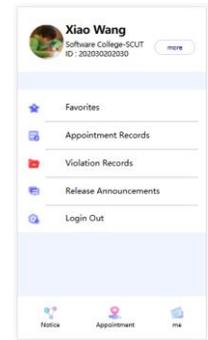

(a) Login Page    (b) Point Page    (c) Person Page

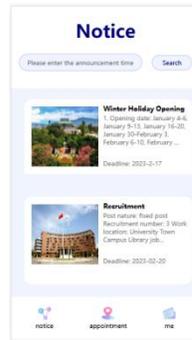 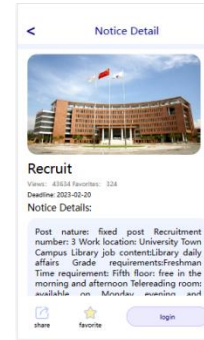 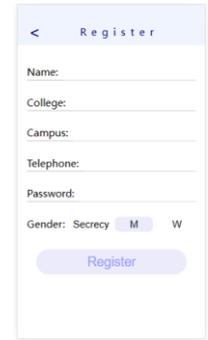

(d) Notice Page    (e) Detail Page    (f) Register Page

Fig. 19. APP interface

### C. Experiments Under Different Lighting and Sparseness of Figures

The test site is the study room on the 2nd floor of the library at the South campus of Xidian University (Xi'an, Shaanxi, China). As in Fig. 20, the test results are dis-played on the Web interface.

In Fig. 20(a), the test time is in the morning with strong light, and in Fig. 20(b), the test time is in the evening with weak light. Fig. 20 (c) shows a sparse distribution of characters compared to Fig. 20(b).

In Fig. 20(a), there are 16 seats, and 2 seats are occupied. The test result show that seats 6 and 12 are occupied. In Fig. 20(b), There are 16 seats in total, and 3 seats are occupied. The test result show that seats 7, 14, and 16 are occupied. In Fig. 20(c), there are 16 seats in total, and 8 seats are occupied. The test result show that seats 1, 3, 5, 8, 9, 11, 12, and 14 are occupied. The results of all three tests are consistent with reality.

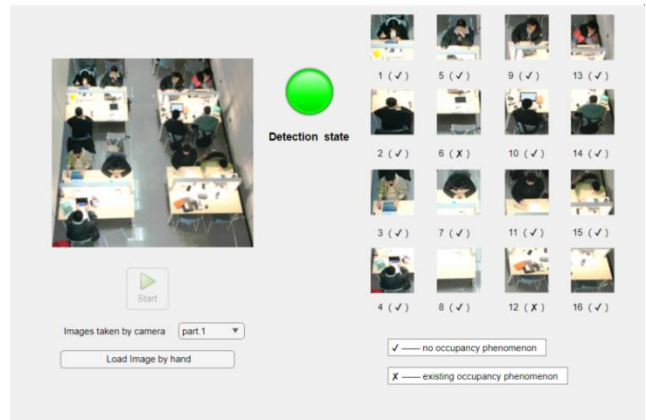

(a) Test result under strong illumination

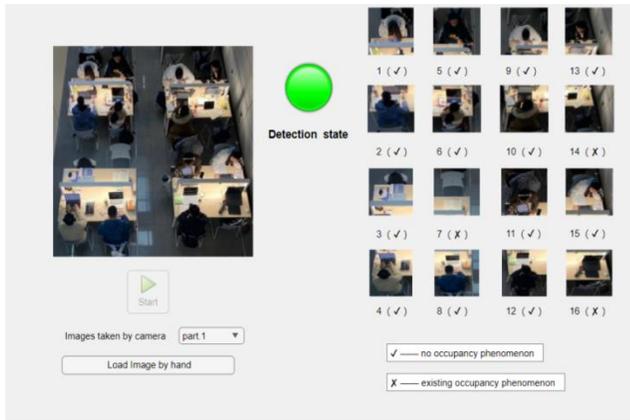

(b) Test result under low light and dense people

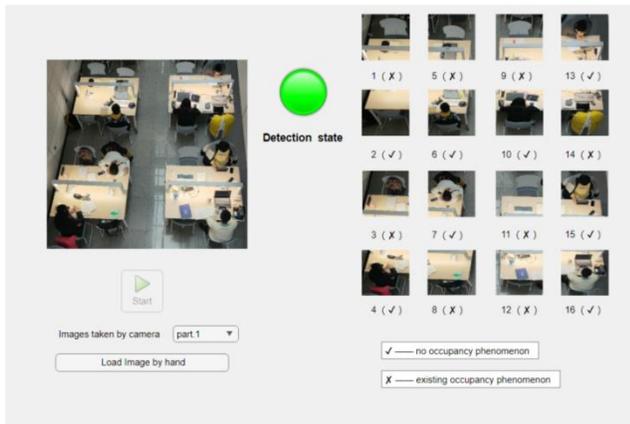

(c) Test result under weak illumination and sparse Fig. s

Fig. 20. Experimental results

## VI. Conclusion

In this paper, we propose a serial dual-channel detection model based on Faster RCNN algorithm for object detection and object classification, and develop a computer vision-based library occupancy detection system. The improved detection model only requires labeling of people and network training, which greatly reduces computational resources and time costs. the Faster RCNN algorithm increases the accuracy of detection. Virtual reality technology provides a massive training set for network training and reduces the cost of human photography. The final results are displayed in a specially designed Web interface and APP, which truly realizes the end-to-end functionality of the system. The system has been experimented in a school library and the final test results show that the system is fully operational and detects the occupancy in the library. At this stage, the model can only cut and detect specific images, which can only contain neatly placed tables and chairs in a top view. The segmentation and detection of images of arbitrary tables and chairs in different views will be a future research direction.

## Acknowledgment

This work was supported by the Fundamental Research Funds for the Central Universities under S202210701123.